\documentclass[10pt]{article} % For LaTeX2e

\usepackage[accepted]{rlj} % Should be uncommented for the camera-ready
\usepackage{amssymb}            % Defines common symbols like \mathbb R
\usepackage{mathtools}          % Extends amsmath, providing common math tools
\usepackage{mathrsfs}           % Enables \mathscr, which can work in cases that \mathcal does not
\usepackage{graphicx}           % For including images
\usepackage{subcaption}         % Allows for the use of subfigures and subcaptions
\usepackage[space]{grffile}     % For spaces in image names
\usepackage{url}                % For displaying URLs
\usepackage{lipsum}             % For placeholder text

\usepackage{amsmath}
\usepackage{amsfonts}
\usepackage{amsthm}

\usepackage{xcolor} % ADDED
\usepackage{amssymb} % ADDED
\usepackage{wasysym} % ADDED
\usepackage{csquotes} % ADDED
\usepackage{array} % ADDED
\usepackage{subcaption} % ADDED
\usepackage[dvipsnames]{xcolor}
\usepackage{verbatim}
\usepackage{booktabs}

\title{Grounding LTL Tasks in Sub-Symbolic RL Environments for Zero-Shot Generalization}

\setrunningtitle{Grounding LTL Tasks in Sub-Symbolic RL Environments for Zero-Shot Generalization}

\author{Matteo Pannacci\textsuperscript{1}, Andrea Fanti\textsuperscript{1}, Elena Umili\textsuperscript{1}, Roberto Capobianco\textsuperscript{2}}

\emails{matteo.pannacci@gmail.com, \ \{fanti,umili\}@diag.uniroma1.it, roberto.capobianco@sony.com}

\affiliations{
$^{1}$\textbf{Sapienza University of Rome}\\
$^{2}$\textbf{Sony AI}\\
}

\contribution{
    We propose a method to jointly learn a multi-task policy and a symbol grounder by exploiting the indirect supervision provided by the environment.
    }
    {
    Previous work has mostly explored learning the symbol grounding while solving single tasks \citep{neural-reward-machines2} and learning multi-task policies with known symbol grounding \citep{ltl2action}. While \citet{encode-ltl-with-rnns} has proposed a method for multi-task learning without symbol grounding knowledge, new advances in transferable task encodings \citep{ltl2action} and symbol grounding \citep{neural-reward-machines2} motivate the study of novel methods that leverage them. We base our method on Neural Reward Machines \citep{neural-reward-machines2}, which solve a Semi-Supervised Symbol Grounding (SSSG) by providing indirect supervision to the symbol grounder based on the structure of the LTL formula.
    }

\contribution{
    We empirically show that our method achieves training and zero-shot transfer performances that are close to the upper bound of knowing the true labeling function and outperform the only existing method that learns multi-task policies without assuming knowledge of the true symbol grounding \citep{encode-ltl-with-rnns}.
    }
    {
    Similarly to existing work in multi-task RL with temporal specifications \citep{ltl2action,encode-ltl-with-rnns,deepltl}, we perform experiments in two visual environments: a discrete Minecraft-like grid-world and a continuous navigation environment. In each environment, we consider two classes of task distributions, one requiring the agent to trigger multiple parallel sequences of events, and another requiring it to complete specific sequences while avoiding a particular event. 
    }

 % Your keywords

\newcommand{\myabstract}{
In this work we address the problem of training a Reinforcement Learning agent to follow multiple temporally-extended instructions expressed in Linear Temporal Logic in sub-symbolic environments. Previous multi-task work has mostly relied on knowledge of the mapping between raw observations and symbols appearing in the formulae. We drop this unrealistic assumption by jointly training a multi-task policy and a symbol grounder with the same experience. The symbol grounder is trained only from raw observations and sparse rewards via Neural Reward Machines in a semi-supervised fashion. Experiments on vision-based environments show that our method achieves performance comparable to using the true symbol grounding and significantly outperforms the only other previous method for multi-task learning that does not assume knowledge of the true symbol grounding.
}
\summary{\myabstract
}

\newtheorem{definition}{Definition}

\newcommand{\LTLnext}{\Circle}

\newcommand{\LTLeventually}{\lozenge}
\newcommand{\LTLuntil}{\mathbin{\mathcal{U}}}

\begin{document}

\makeCover  % Create the cover page
\maketitle  % Make the title section

\begin{abstract}
\myabstract
\end{abstract}

\section{Introduction}

Creating agents that can follow a vast set of instructions rather than solving only a specific task has been a central goal in Reinforcement Learning (RL) \citep{transfer_learning_RL_survey}. Formal languages such as Linear Temporal Logic (LTL) \citep{LTL} are widely adopted to specify such tasks because of their unambiguous semantics and compositional syntax \citep{restraining-bolts}. LTL instructions are built from atomic propositions, or \emph{symbols}, representing relevant features of the environment. These propositions are combined through logical and temporal operators to express temporally extended tasks. For example, symbols in an office environment might represent being at the coffee machine, a mail pickup site or a specific desk. A basic temporally-extended task might then be (informally), ``pick up the mail and coffee in this order and bring them to my desk''.
However, if the agent is equipped with only raw sensory equipment, such as a video camera, it must first identify the relevant office locations from its observations before reasoning about the instruction it received. In general, interpreting an LTL instruction requires knowledge of both LTL semantics and a mapping from observations to symbolic interpretations, known as \emph{symbol grounding}.

We can expect many real-world domains like the example above to not provide access to the true symbol grounding. These are known as \emph{sub-symbolic} environments. Learning a multi-task policy in sub-symbolic environments requires the agent to simultaneously learn: (i) a policy that captures the semantics of LTL operators; (ii) a symbol grounder connecting observations to propositions. While learning multi-task policies using the true symbol grounder has been explored in previous works, the more realistic setting is rarely treated due to the compounding complexity of the two learning problems. To the best of our knowledge, only \citet{encode-ltl-with-rnns} treat this combined problem with an approach that is substantially different from other multi-task work \citep{ltl2action,deepltl}. As shown by our experiments, this method is vulnerable to long formulae and performs significantly worse than the upper bound of a known symbol grounding.

% Learning a multi-task policy in such environments requires the agent to simultaneously learn: (i) how to interpret LTL specifications by learning the semantics of the LTL operators; (ii) how to act in the environment to accomplish the requested task; and (iii) how to ground atomic propositions appearing in the tasks into raw observations. While previous work has extensively studied the first two capabilities assuming access to the true symbol grounding, the more realistic setting in which all three must be learned jointly has received little attention because of their strong interdependence.

Here we address these two challenges together and propose a method to learn multi-task policies that generalize to unseen instructions in sub-symbolic environments. Our method builds on Neural Reward Machines (NRMs) \citep{neuralrewardmachines}, which were previously only applicable to single tasks in isolation, to jointly learn a multi-task policy and a symbol grounder using the same experience. NRMs frame the problem as a Semi-Supervised Symbol Grounding (SSSG) task to provide indirect supervision to the symbol grounder based on the structure of the LTL formulae. The resulting agent can follow LTL instructions by tracking the task progression directly from raw observations.
Our experiments show that, while NRMs are known to struggle with sparse rewards in single-task settings, training on multiple tasks provides a sufficient learning signal for NRMs to correctly infer the symbol grounding.
We use the state-of-the-art method LTL2Action \citep{ltl2action} to condition the policy on task progression information. Our approach is however not tied to the specifics of LTL2Action and can be combined with any transferable task progression encoding.

We study the convergence and zero-shot generalization power of our method on a Minecraft-like discrete environment and a continuous navigation environment. Our method shows little to no loss in performance with respect to using the true labeling function. Our method also improves significantly over \citet{encode-ltl-with-rnns}, the only previous method for multi-task RL in sub-symbolic environments.

The main contributions of our work are then:
\begin{itemize}
    \item a neurosymbolic architecture designed for jointly learning a multi-task policy and a symbol grounder module\footnote{We provide the source code at \hyperlink{https://github.com/KRLGroup/SymGroundMultiTask}{https://github.com/KRLGroup/SymGroundMultiTask}}, which is trained using only the symbolic specification of the task (the LTL formula) and trajectories collected in the environment, without access to the labeling function;
    \item an empirical evaluation in visual environments of both training and zero-shot-transfer performances of the proposed method against the upper bound of knowing the true labeling function and the only multi-task baseline with unknown symbol grounding.
\end{itemize}

\section{Background}
\paragraph{Notation}
We consider \textit{sequential} data in both symbolic and sub-symbolic form. Symbolic sequences, also called \textit{traces}, consist of symbols $\sigma$ drawn from a finite alphabet $\Sigma$. Sequences are denoted in bold, e.g. $\boldsymbol{\sigma} = (\sigma^{(1)}, \sigma^{(2)}, \ldots)$. We denote the empty sequence with $\epsilon$.
Each symbolic variable can be grounded either categorically or probabilistically. In the categorical case, each element is assigned a symbol $\sigma^{(i)} \in \Sigma$. In the probabilistic case, it is associated with a distribution over $\Sigma$, represented by a vector $\tilde{\sigma}^{(i)} \in \Delta(\Sigma)$, where
\[
\Delta(\Sigma) = \left\{ \tilde{\sigma} \in \mathbb{R}^{|\Sigma|} \,\middle|\, \tilde{\sigma}_j \geq 0,\ \sum_{j=1}^{|\Sigma|} \tilde{\sigma}_j = 1 \right\}.
\]
Superscripts denote time steps and subscripts vector components; for example, $\tilde{\sigma}^{(i)}_j$ is the $j$-th component at time step $i$. Accordingly, $\boldsymbol{\sigma}$ denotes a categorically grounded sequence, while $\boldsymbol{\tilde{\sigma}}$ denotes a probabilistically grounded one.
Finally, when approximating a function $f$ with a parametrized model, we denote by $f$ the ground-truth function and by $f_\theta$ its learnable approximation.

\paragraph{LTL}
Linear Temporal Logic (LTL) \citep{LTL} extends propositional logic with temporal operators to specify properties over time. Given a set of propositions $P$, the syntax of an LTL formula $\varphi$ is
\begin{equation} \label{eq:LTL_syntax}
    \varphi := \top \mid \bot \mid p \mid \lnot\varphi \mid \varphi_1 \wedge \varphi_2 \mid \mathsf{X}\varphi \mid \varphi_1 \mathsf{U} \varphi_2,
\end{equation}
where $p \in P$. Here, $\top$ and $\bot$ denote truth values, while $\mathsf{X}$ (next) and $\mathsf{U}$ (until) are temporal operators. Derived operators include $\mathsf{F}\varphi \equiv \top \mathsf{U} \varphi$ (eventually) and $\mathsf{G}\varphi \equiv \lnot \mathsf{F}(\lnot\varphi)$ (globally). We refer to \citet{LTL} for formal semantics.
An LTL formula is interpreted over an infinite trace $\boldsymbol{\sigma} = (\sigma^{(0)}, \sigma^{(1)}, \ldots)$, where $\sigma^{(t)} \in 2^{P}$ denotes the \textit{interpretation} at time $t$, namely the set of propositions that are true at that time.
We write $\boldsymbol{\sigma} \vDash \varphi$ when the trace $\boldsymbol{\sigma}$ satisfies $\varphi$. When propositions in $P$ are mutually exclusive, exactly one proposition holds at each time step, i.e., $\sigma^{(t)} \in P$; this is the so-called \emph{Declare assumption} \citep{declare_assumptiopn}, which we adopt in this work. This assumption lets us formulate the symbol grounding problem as a classification problem with $|P|$ classes. This is without loss of generality, since subsets of propositions can be encoded as propositions belonging to a new alphabet $P' = 2^{P}$. Moreover, it has no impact on the baseline considered in our experiments \citep{encode-ltl-with-rnns}, whose architecture directly learns latent representations of observation sequences and formulas rather than explicitly predicting interpretations. Finally, we focus on \emph{co-safe} LTL formulae \citep{model-checking-for-safety}, that is, formulae whose satisfaction can be determined from a finite prefix of the trace.

\paragraph{LTL Progression}
\begin{comment}
The progression algorithm \citep{planning-for-temporally-extended-goals,ltlprogression} is a systematic procedure that takes an LTL formula and a truth assignment as input and produces a new formula representing those aspects of the original instructions that must still be addressed, making it useful for tasks such as runtime verification and planning. Formally, given an LTL formula $\varphi$ and a truth assignment $\sigma_i$ over $\mathcal{P}$, the progression of $\varphi$ through $\sigma_i$, denoted by $prog(\sigma_i,\varphi)$, is defined as follows:
\begin{align*}
    prog(\sigma_i,p) = & \begin{cases}
        \top  & \text{if } p \in \sigma_i \\
        \bot & \text{if } p \notin \sigma_i
    \end{cases} \quad \text{where } p \in \mathcal{P} \\
    prog(\sigma_i, \neg \varphi) = & \; \neg prog(\sigma_i, \varphi) \\
    prog(\sigma_i, \varphi_1 \wedge \varphi_2) = & \; prog(\sigma_i, \varphi_1) \wedge prog(\sigma_i, \varphi_2) \\
    prog(\sigma_i, \LTLnext \varphi) = & \; \varphi \\
    prog(\sigma_i, \varphi_1 \LTLuntil \varphi_2) = & \; prog(\sigma_i, \varphi_2) \; \vee \\
    & (prog(\sigma_i, \varphi_1) \wedge (\varphi_1 \LTLuntil \varphi_2))
\end{align*}
\end{comment}
The progression algorithm \citep{planning-for-temporally-extended-goals,ltlprogression} rewrites an LTL formula according to the current interpretation, yielding a formula that captures the remaining temporal requirements. Given an LTL formula $\varphi$ and the interpretation $\sigma^{(i)} \in 2^{P}$ at time $i$, the progression of $\varphi$ through $\sigma^{(i)}$, denoted by $prog(\sigma^{(i)}, \varphi)$, is defined as:
\begin{align*}
    prog(\sigma^{(i)}, p) = & 
    \begin{cases}
        \top  & \text{if } p \in \sigma^{(i)} \\
        \bot  & \text{otherwise}
    \end{cases}
    \qquad p \in P \\
    prog(\sigma^{(i)}, \lnot \varphi) = & \; \lnot prog(\sigma^{(i)}, \varphi) \\
    prog(\sigma^{(i)}, \varphi_1 \wedge \varphi_2) = & \; prog(\sigma^{(i)}, \varphi_1) \wedge prog(\sigma^{(i)}, \varphi_2) \\
    prog(\sigma^{(i)}, \mathsf{X}\varphi) = & \; \varphi \\
    prog(\sigma^{(i)}, \varphi_1 \mathsf{U} \varphi_2) = & \; prog(\sigma^{(i)}, \varphi_2) \;\vee \bigl(prog(\sigma^{(i)}, \varphi_1) \wedge (\varphi_1 \mathsf{U} \varphi_2)\bigr).
\end{align*}

\paragraph{Moore Machines}
A Moore machine $M = (P, Q, O, q_0, \delta, \lambda)$ consists of a finite input alphabet $P$, states $Q$, output symbols $O$, an initial state $q_0 \in Q$, a transition function $\delta: Q \times P \to Q$, and an output function $\lambda: Q \to O$.  
For a finite input trace $\boldsymbol{\sigma} = (\sigma^{(0)}, \dots, \sigma^{(n)})$, the extended transition $\delta^*: Q \times P^* \to Q$ is defined recursively as $\delta^*(q, \epsilon) = q$ and $\delta^*(q, \sigma_1 \dots \sigma_n) = \delta(\delta^*(q, \sigma_1 \dots \sigma_{n-1}), \sigma_n)$. The output sequence produced by $M$ is $
\lambda(\boldsymbol{\sigma}) = (\lambda(q_0), \lambda(q_1), \dots, \lambda(q_n))$, with $q_t = \delta^*(q_0, \sigma^{(0)}\dots\sigma^{(t-1)}), \ t \ge 1.
$

\paragraph{Non-Markovian Reinforcement Learning}
In reinforcement learning, agent-environment interaction is commonly modeled as a Markov Decision Process (MDP) \citep{sutton}, defined by the tuple $(S, A, t, r, \gamma)$, where $S$ is the set of states, $A$ the set of actions, $t : S \times A \times S \rightarrow [0,1]$ the transition function, $r : S \times A \rightarrow \mathbb{R}$ the reward function, and $\gamma \in [0,1]$ the discount factor. A \emph{policy} $\pi : S \rightarrow A$ maps states to actions, while the \emph{value function} $V^\pi(s)$ denotes the expected discounted return obtained by following $\pi$ from state $s$. The goal of the RL agent is to learn the optimal policy $\pi^*$ providing maximum value.
Standard MDPs assume Markovian transitions and rewards, i.e., dependence only on the current state. However, many real-world problems violate this assumption \citep{littman2017environment}. In a non-Markovian decision process, the reward function $r : (S \times A)^* \rightarrow \mathbb{R}$, the transition function $t : (S \times A)^* \times S \rightarrow [0,1]$, or both may depend on the interaction history. In this work, we focus on Non-Markovian Reward Decision Processes (NMRDPs) \citep{non-markow-rewards1996}, where non-Markovianity arises solely from the reward function.
Learning optimal policies in NMRDPs is challenging, as rewards may depend on the entire sequence of past state--action pairs, making standard RL methods inapplicable. To address this issue, many approaches construct an augmented Markovian state representation by monitoring the goal through a labeling function $L : S \rightarrow P$, which maps environment states to propositional symbols. The resulting labeled traces are used to track task progress, for example via LTL formula progression \citep{multi-task-rl,ltl2action} or equivalent automata-based representations such as Moore machines \citep{reward-machines-ltl,restraining-bolts}.

\begin{figure*}[t]
    \centering
    \begin{subfigure}[t]{0.35\linewidth}
      \centering
      \includegraphics[width=\linewidth]{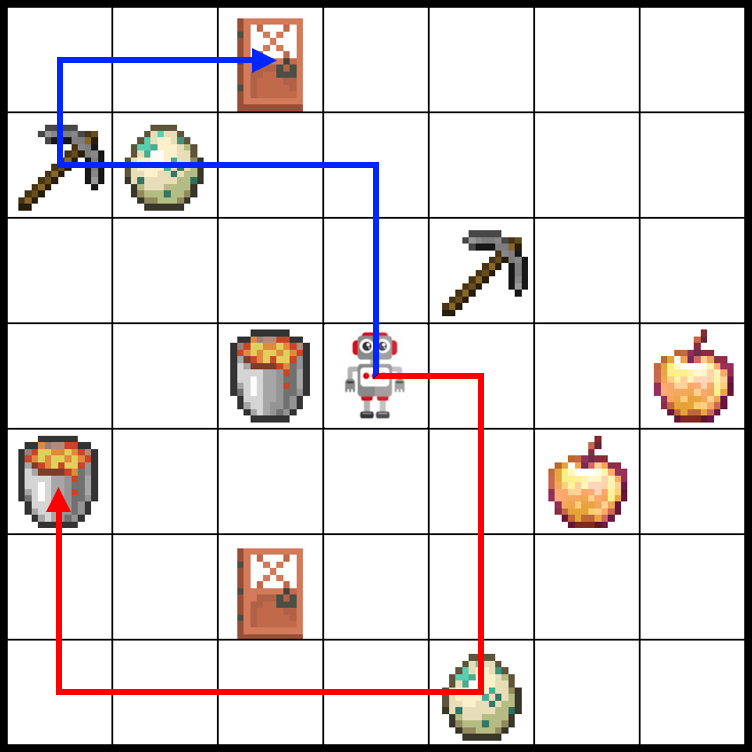}
      \caption{Minecraft-Like Environment}
      \label{fig:example_environment_grid}
    \end{subfigure}
    \hspace{0.05\textwidth}
    \begin{subfigure}[t]{0.41\linewidth}
      \centering
      \includegraphics[width=\linewidth]{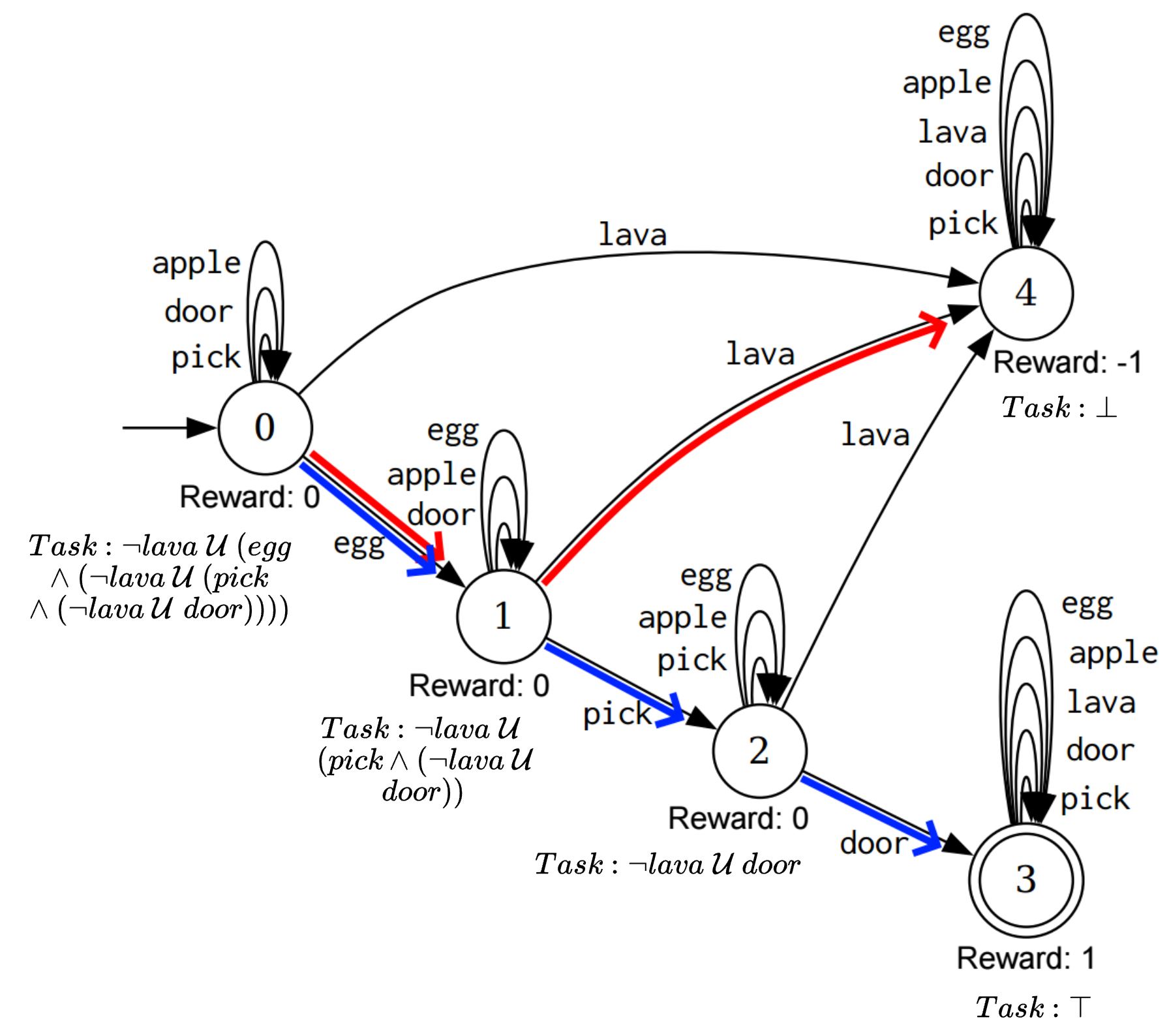}
      \caption{Task's Moore machine}
      \label{fig:example_environment_automaton}
    \end{subfigure}
    \caption{Environment visualization and Moore machine corresponding to the task $\neg lava \LTLuntil (egg \wedge (\neg lava \LTLuntil (pick \wedge (\neg lava \LTLuntil door))))$, which is the (co-safe) task of following the sequence $egg \rightarrow pick \rightarrow door$ without traversing $lava$ in the meanwhile. In blue is represented a trace of execution leading to success, while in red a trace of execution leading to failure.}
    \label{fig:example_automaton}
\end{figure*}

\paragraph{Example: Image-Based Minecraft-Like Environment} \label{par:example}
Consider the grid-world in Figure \ref{fig:example_environment_grid} containing a pickaxe, an egg, an apple, a door, and a lava cell. The task is to reach the egg, the pickaxe, and the door in this order while avoiding lava, expressed as the co-safe LTL formula $
\neg lava \ \mathsf{U} \ (egg \wedge (\neg lava \ \mathsf{U} \ (pick \wedge (\neg lava \ \mathsf{U} \ door))))$.
This formula can be converted into the Moore machine $M$ shown in Figure \ref{fig:example_environment_automaton}, with $P = \{\text{pickaxe, apple, door, egg, lava}\}$. Each symbol is true when the agent is on the corresponding cell and false otherwise. The agent observes at each step the current state $s \in S$, given as an image of the grid-world. This representation is non-Markovian, as the current image reveals only the agent's present location and not which items or cells have been visited previously, nor the current state of the Moore machine.

\section{Related Work}

\paragraph{Reinforcement Learning with Temporal Specifications}
Temporal logic formalisms are often used in reinforcement learning with non-Markovian rewards to define temporally-extended goals and constraints. Popular choices are Linear Temporal Logic (LTL) \citep{LTL} and its variant evaluated on finite traces, LTL$_f$ \citep{ltlf13}. These make it possible to recover the Markovian property by conditioning the policy on the state of an automaton constructed from the formula. Examples include Restraining Bolts \citep{restraining-bolts,restraining-bolts2} and Reward Machines \citep{rewardmachines,reward-machines-ltl,reward-machines-ltl2}. These methods can only learn a single task or a fixed set of tasks and assume perfect knowledge of the labeling function.

\paragraph{Reinforcement Learning without Perfect Labeling}
Several methods have considered a noisy or imperfect labeling function \citep{noisysymbolicabstraction,rl-based-temporal-logic-control,joint-learning-of-reward-machines-and-policies} but without attempting to learn the symbol grounder itself.
\citet{neuralrewardmachines} introduced Neural Reward Machines (NRMs), which allow to learn the symbol grounder directly from sequences of observations and rewards without any access to the true labeling function. If the task dynamics is unknown, NRMs can also learn the reward machine’s transition and reward functions \citep{neural-reward-machines2}. These can still only learn single tasks or fixed sets of tasks.

\paragraph{Multi-Task Reinforcement Learning}
A popular approach to learning multi-task policies is decomposing specifications into subtasks solved individually, both with LTL formulae \citep{systematic-generalization-through-ltl,in-a-nutshell-the-human-asked-for-this,the-logical-options-framework,ltl-transfer} and with other specifications \citep{modular-multitask-rl-with-policy-sketches,neural-task-programming,zero-shot-task-generalization-with-multi-task-drl,hierarchical-reiforcement-learning-for-zero-shot-generalization}. However, despite enabling generalization to new tasks, decomposition can result in suboptimal solutions due to uncoordinated subtask policies.
\citet{multi-task-rl} applied LTL progression \citep{ltlprogression} to the problem, to dynamically update the LTL formula as the agent progresses in the task. Their algorithm provably converges to the optimal multi-task policy but does not generalize to unseen tasks.
LTL2Action \citep{ltl2action} combines LTL progression and Graph Neural Networks to encode the progressed formula, enabling generalization to new tasks. They also employ environment-agnostic pre-training of the formula encoder, boosting performance significantly.
\citet{transfer-learning-ltl} explored a non-learned transferable encoding of reward machine states based on semantic similarity, obtaining mixed results.
% \notes{Finally, \citep{transfer-learning-ltl} proposed an approach in which a transferable task state representation is computed through the semantic similarity (obtained through a kernel function) between the task state and the state of a set of reference tasks, combined into a vector representation. This representation is then combined with the environment state to make the decision process Markovian. The advantage of this method is that, unlike most other approaches, these task representations do not require specialized architectures or additional learning during training [remove?]}.
\citet{deepltl} introduced a representation for LTL formulae based on reach-avoid sequences of truth assignments, allowing to learn multi-task policies on infinite-horizon tasks.
%\citep{encode-ltl-with-rnns} developed a method that assembles a task-specific recurrent network by composing learned sub-networks modules corresponding to the syntactic elements of the task formula. The resulting network takes as input the environment's observations and returns the action to execute, as such the structure is strongly environment-specific but it can generalize to unseen tasks.
\citet{encode-ltl-with-rnns} is the only existing work that learns multi-task policies that can generalize to new tasks and doesn't require knowledge of the symbol grounding. This method combines learned sub-networks into task-specific networks by following the syntactic structure of the LTL formula. More specifically, the method trains a recurrent network module for each proposition and LTL operator and then combines them into a tree where each node feeds the previous state of its recurrent network to its children and the next state to its parent. This is a radically different approach with respect to more recent multi-task methods, which instead use single encoder modules to directly condition the policy on task progression information represented as a sequence or graph \citep{ltl2action,deepltl}. Nonetheless, we compare to this method in our experiments since it is the only one tackling our same problem setting of learning multi-task policies with unknown symbol grounding.

% In this work, we adopt the same setting and extend the framework of \citep{ltl2action} by integrating Neural Reward Machines to jointly learn the grounding function along with the policy. This integration preserves the original modularity of the framework, keeping the LTL module environment-agnostic while retaining the ability to pretrain and reuse the learned semantic representations of LTL formulae.

\section{Method}

\subsection{Problem Setting and Objectives}
We consider the problem of generalizing across temporal tasks expressed in LTL in sub-symbolic environments where the labeling function is unknown. Let $M=(S,A,t,\gamma)$ be a reward-free MDP, let $\varphi$ be a co-safe LTL formula over a set of propositions $P$, and let $L:S\rightarrow P$ be a labeling function. For each task specification $\varphi$, we define the corresponding NMRDP $M_\varphi=(S,A,t,r_\varphi,\gamma)$, where $r_\varphi$ provides a three-valued non-Markovian reward signal:
\begin{equation} \label{eq:reward} 
r_{\varphi}(s_0, a_0, \cdots, s_t, a_t) = \begin{cases} 1 & \text{if } \langle L(s_0), \cdots, L(s_t) \rangle \text{ is} \\ & \text{a \textit{good prefix} for } \varphi \\ -1 & \text{if } \langle L(s_0), \cdots, L(s_t) \rangle \text{ is} \\ & \text{a \textit{bad prefix} for } \varphi \\ 0 & \text{otherwise} \end{cases} 
\end{equation}
Where a \emph{good prefix} is a trace that guarantees satisfaction and a \emph{bad prefix} is a trace that guarantees falsification \citep{model-checking-for-safety}. In these cases, we also terminate the episode. This corresponds to the $LTL_3$ semantics \citep{runtime-verification-for-ltl}, providing minimal feedback sufficient to convey the task. Denser rewards could be derived via subtask potentials \citep{rewardmachines,neuralrewardmachines}, but are omitted here to test a general scenario.
Let $\Phi_{train}$ and $\Phi_{test}$ be the training and test formula sets, with corresponding environment sets $E_{train} = \{M_\varphi \mid \varphi \in \Phi_{train}\}$ and $E_{test} = \{M_\varphi \mid \varphi \in \Phi_{test}\}$. Transfer occurs to environments with different temporal tasks but identical perception complexity and dynamics.  
We assume access to the initial task $\varphi$ during training or testing on $M_\varphi$ but not to $L$. The goal is to leverage the syntactic and semantic structure of training tasks and trajectories to learn a non-Markovian policy $\pi: H \rightarrow A$, with $H$ the set of all trajectories $\langle s_0, a_0, \dots, s_t, a_t \rangle$, that can transfer to unseen tasks without retraining.

\begin{figure*}[t]
  \centering
  \includegraphics[width=\textwidth]{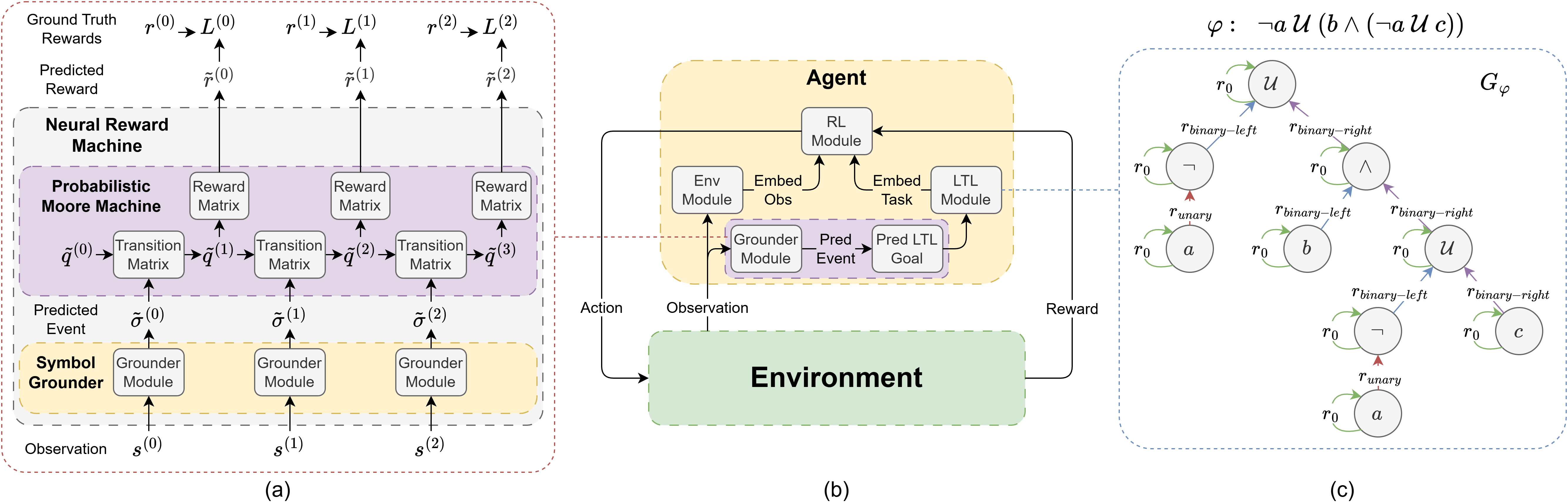}
  \caption{
  (a) Unfolded computational graph of the grounder training through the LTL task's Neural Reward Machine, employing backpropagation through time. $\tilde{q}^{(0)}$ is the initial state of the NRM and $\tilde{q}^{(t)}$ denotes the predicted state at the start of step $t$.
  (b) Overview of the RL framework. %The env module embeds the current observation received from the environment. The grounder module approximates the labeling function, predicting the event associated with the current observation. Based on these predicted events and the initial LTL goal, the agent maintains its own estimation of the progressed LTL formula over time. This estimated progressed goal is embedded by the LTL module. Finally, the RL module combines the embeddings produced by the env and LTL modules to generate the policy’s probability distribution over the action space.
  %While the agent relies on its internally estimated progression of the LTL formula to condition its policy, the reward signal is computed using the true progression of the LTL goal, as defined by the real labeling function (unknown to the agent).
  (c) The LTL goal represented as a formula and as the corresponding graph derived from its AST. 
  }
  \label{fig:framework}
\end{figure*}

\subsection{Symbolic Multi-Task RL over LTL}
\begin{comment}
For now, we assume that the labeling function $L$ is known and introduce policy learning in a symbolic setting. Section \ref{sec:sym_grounder} will discuss in more detail the learning of this function.

Among symbolic methods, those based on \emph{progression} \citep{multi-task-rl,ltl2action} have proven most promising for multi-task RL over temporal goals. \citep{multi-task-rl} first proposed using the progression algorithm to update the LTL goal provided to the agent during episode execution. In this approach, the formula is updated at each step $t$ using the current truth assignment $L(s\textsuperscript{($t$)}, a\textsuperscript{($t$)})$ and simplified to remove satisfied parts while retaining those yet to be satisfied. It was proved that, the progressed formula suffices to construct a Markovian representation of a trajectory $h = \langle s\textsuperscript{($0$)}, a\textsuperscript{($0$)}, \dots, s\textsuperscript{($t$)}, a\textsuperscript{($t$)} \rangle$, as the state space $S \times \mathcal{L}_{\mathrm{LTL}}(P)$, where $P$ is the set of atomic propositions and $\mathcal{L}_{\mathrm{LTL}}(P)$ is the set of all possible LTL formulae over $P$ \citep{multi-task-rl}. More formally:
\begin{align*}
\pi(s^{(0)}, a^{(0)}, \dots, s^{(t)}, a^{(t)}) &= \pi(s^{(t)}, \varphi^{(t)}), \\
\varphi^{(t)} &= prog(L(s^{(t)}), \varphi^{(t-1)}), \\
\varphi^{(0)} &= \varphi.
\end{align*}
\end{comment}

For clarity, we first assume that the labeling function $L$ is known and introduce policy learning in the symbolic setting. Learning $L$ itself is addressed in Section \ref{sec:sym_grounder}.

Among symbolic methods, \emph{progression}-based approaches \citep{multi-task-rl,ltl2action} have proven particularly effective for multi-task RL with temporal goals. \citet{multi-task-rl} proposed updating the LTL goal at each step $t$ using the current truth assignment $L(s^{(t)})$ through LTL progression, simplifying satisfied parts while keeping those yet to be satisfied. 
Augmenting the state with the progressed formula provides a Markovian state representation: a trajectory $h = \langle s^{(0)}, a^{(0)}, \dots, s^{(t)}, a^{(t)} \rangle$ can be represented by the augmented state $s' = (s^{(t)}, \varphi^{(t)}) \in S \times \mathcal{L}_{\mathrm{LTL}}(P)$, where $P$ is the set of propositions and $\mathcal{L}_{\mathrm{LTL}}(P)$ is the set of all LTL formulae over $P$ \citep{multi-task-rl}.
Formally $\pi(s^{(0)}, a^{(0)}, \dots, s^{(t)}, a^{(t)}) = \pi(s^{(t)}, \varphi^{(t)})$, with
\begin{align*}
\varphi^{(t)} =
\begin{cases}
\varphi & t=0 \\
prog(L(s^{(t)}), \varphi^{(t-1)}) & t>0
\end{cases},
\end{align*}

\paragraph{Example: LTL monitoring in the Minecraft-Like Environment}
In Figure \ref{fig:example_automaton}, the original task $ \varphi = \varphi^{(0)} = \neg lava \LTLuntil (egg \wedge (\neg lava \LTLuntil (pick \wedge (\neg lava \LTLuntil door))))$ progresses whenever the agent reaches a relevant item. For example, reaching the egg (the first target) moves the Moore machine from state 0 to state 1, equivalent to progressing the formula from $ \varphi^{(0)} =\neg lava \LTLuntil (egg \wedge (\neg lava \LTLuntil (pick \wedge (\neg lava \LTLuntil door))))$ to $ \varphi^{(1)} = \neg lava \LTLuntil (pick \wedge (\neg lava \LTLuntil door))$. Figure \ref{fig:example_environment_automaton} shows, for each automaton state, the corresponding progressed formula and the reward received upon reaching it.

\subsection{System Overview}
Building on these theoretical insights, we design the architecture shown in Figure~\ref{fig:framework}b. It consists of four modules, each implemented as a neural network:

\begin{itemize}
    \item The \textbf{grounder module}: approximates the unknown labeling function $L$ by mapping raw observations to symbols, $L_\theta: S \rightarrow \Delta(P)$.
    \item The \textbf{environment feature module}: encodes observations into feature vectors, $f^{img}_\theta: S \rightarrow \mathbb{R}^n$, where $n$ denotes the feature dimension.    
    \item The \textbf{LTL module}: encodes the progressed task formula $\varphi^{(t)}$, $t \geq 0$, as a feature vector, $f^{LTL}_\theta: \mathcal{L}_{\mathrm{LTL}}(P) \rightarrow \mathbb{R}^m$, where $m$ is the dimension of the LTL feature space.
    \item The \textbf{RL module}: receives the concatenated feature vectors $f^{img}_\theta(s^{(t)})$ and $f^{LTL}_\theta(\varphi^{(t)})$ as input and learns both a policy $\pi_\theta: \mathbb{R}^{n+m} \rightarrow \Delta(A)$ and a value function $V_\theta: \mathbb{R}^{n+m} \rightarrow \mathbb{R}$.
\end{itemize}

At each step, the current observation $s^{(t)}$ is processed by both the environment and grounder modules. The environment module encodes the feature vector $f^{img}_\theta(s^{(t)})$, while the grounder module predicts the most likely symbol $p = \arg\max L_\theta(s^{(t)})$. This symbol is used to progress the task formula from $\varphi^{(t-1)}$ to $\varphi^{(t)}$, which the LTL module encodes as the feature vector $f^{LTL}_\theta(\varphi^{(t)})$. The two feature vectors are then concatenated and processed by the RL module to select the next action.
This pipeline closely resembles that of LTL2Action \citep{ltl2action}, with one key difference: in our system, \textit{all} modules are trainable, whereas in LTL2Action the grounding function is known a priori. Our grounder module is trained alongside the other modules and is imperfect, especially early in training. As a result, errors in symbol grounding propagate to the progressed formula $\varphi\textsuperscript{($t$)}$ on which the policy relies when selecting actions. The RL and environment modules are trained end-to-end with the RL loss, while the grounder and LTL modules have separate objectives described in the next sections.

\subsection{Grounder Training} \label{sec:sym_grounder}
The \textit{grounder module} is a neural network classifier that cannot be trained via standard supervised learning, as explicit labels for each observation are unavailable, nor fully end-to-end with Deep RL, since LTL progression can't be applied with probabilistic symbol predictions. Instead, the agent learns the grounding indirectly from the environment through sequences of rewards and observed states. This setting falls under \textit{Semi-Supervised Symbol Grounding} (SSSG) \citep{neural-probabilistic-logic-programming,logic-tensor-networks,grounding-ltlf-specifications-in-image-sequences}, where the goal is to map raw observations to symbolic representations using limited or indirect supervision. In this work, we leverage Neural Reward Machines (NRMs) \citep{neuralrewardmachines} to perform semi-supervised knowledge injection for learning the labeling function. Crucially, instead of only using NRMs for single isolated tasks as in previous work, we apply them to the more complex multi-task setting. This allows us to drop the assumption of known symbol grounding, prevalent in the multi-task literature, and mitigate known problems of NRMs in sparse reward settings \citep{neuralrewardmachines}, as shown later in our experiments.

\paragraph{Neural Reward Machines}
NRMs are a probabilistic extension of Reward Machines (RMs) \citep{rewardmachines} that model non-Markovian rewards via an automata-based structure, while incorporating uncertainty in transitions, rewards, and symbol grounding. Formally, an NRM is a Probabilistic Moore machine with a probabilistic symbol grounding function mapping environment states to symbols and interpreting output symbols as rewards.

% \begin{definition}
A Neural Reward Machine (NRM) is a tuple $\langle S, P, Q, R, \mu, \mathcal{T}, \mathcal{R}, L \rangle$, where $S$ is the set of environment states, $P$ a finite symbol set, $Q$ the machine states, $R$ the rewards, $\mu \in \Delta(Q)$ the initial state distribution, $\mathcal{T}[p,q,q']$ the transition probability from $q$ to $q'$ on symbol $p$, $\mathcal{R}[q,r]$ the reward probability for $r$ in state $q$, and $L: S \rightarrow \Delta(P)$ the probabilistic grounding function.
% \end{definition}

Both the machine state $\tilde{q}^{(t)} \in \Delta(Q)$ and output $\tilde{r}^{(t)} \in \Delta(R)$ are stochastic, and NRMs generalize further by allowing the input $\tilde{p}^{(t)} \in \Delta(P)$ to be probabilistic. Following \citet{neuralrewardmachines}, the parametrized model can be implemented as a neural network:
\begin{equation} \label{eq:transition_rnn}
\begin{alignedat}{2}
\mu &= \text{softmax}(\theta_\mu / \tau), \quad
\mathcal{T} = \text{softmax}(\theta_\mathcal{T} / \tau) \\
\mathcal{R} &= \text{softmax}(\theta_\mathcal{R} / \tau), \quad
\tilde{p}^{(t)} = L_\theta(s^{(t)}) \\
\tilde{q}^{(t)} &=
\begin{cases}
\mu & t=0 \\
\sum_j \tilde{p}^{(t)}[j]\,
(\tilde{q}^{(t-1)} \cdot \mathcal{T}[j]) & t>0
\end{cases}, \quad
\tilde{r}^{(t)} = \tilde{q}^{(t)} \cdot \mathcal{R}
\end{alignedat}
\end{equation}
Figure \ref{fig:framework}a illustrates the NRM. The grounding function $L_\theta$ can be any neural network outputting a probability over $P$. The parameters $\theta_\mu$, $\theta_\mathcal{T}$, and $\theta_\mathcal{R}$ define the initial state, task transitions, and rewards. In our setting, we initialize these parameters to encode the task $\varphi$ for each environment $M_\varphi$, and train only the grounder $L_\theta$ using observed sequences of states and rewards.

\subsubsection{From LTL formulae to Moore Machines}
So far, we expressed tasks as LTL formulae interpreted under $LTL_3$ semantics \citep{runtime-verification-for-ltl}. We now derive an equivalent Moore machine representation whose outputs indicate whether the current trace has permanently violated ($-1$), can still evolve ($0$), or has permanently satisfied ($+1$) the specification.

Co-safe LTL formulae can be translated into DFAs recognizing their \textit{good prefixes} \citep{model-checking-for-safety}. To compute the reward function, we must also identify the \textit{bad prefixes}, namely sequences that cannot be extended into a satisfying trace. These correspond to non-final states from which no accepting state is reachable, commonly referred to as \textit{dead states}. They can be computed as the complement of \textit{live states}, i.e., states from which a final state is reachable.

Let $\langle Q, P, \delta, q_0, F \rangle$ denote the DFA recognizing the good prefixes of $\varphi$, with $F$ and $D$ denoting its final and dead states respectively. The corresponding Moore machine is $\langle P, Q, O, q_0, \delta, \lambda \rangle$ where $O = \{1, 0, -1\}$ and
\begin{align*}
    \lambda(q) = \begin{cases}
        1 & q \in F \\
        -1 & q \in D \\
        0 & \text{otherwise}
    \end{cases}.
\end{align*}

\subsubsection{Learning with NRMs}
NRMs provide a differentiable implementation of automata, enabling backpropagation to the grounder module. Unlike \citet{neuralrewardmachines}, we employ multiple NRMs, each representing a different training task through its own Moore machine while sharing the same grounder.
Grounder training consists of two phases: data collection and update. During collection, RL interactions are stored as state and reward sequences $\boldsymbol{s}=\langle s^{(0)},\dots,s^{(t)}\rangle$ and $\boldsymbol{r}=\langle r^{(0)},\dots,r^{(t)}\rangle$, together with the task formulae $\varphi$, allowing data collection  without additional simulations.

% The grounder predicts symbols for LTL progression by selecting the most probable symbol, allowing experience collection without additional simulations.

As noted in \citet{neuralrewardmachines}, episodes with only zero rewards provide little supervision and exacerbate the \emph{reasoning shortcuts} problem \citep{RS_survey}, which is common in NeSy predictors for SSSG: when symbolic knowledge is underspecified, multiple symbol mappings may satisfy the same structure, leading to unintended solutions. While \citet{neuralrewardmachines} addresses this issue using dense, potential-based rewards, in our training setup we eliminate the reliance on dense rewards by exploiting the structure induced by multiple training tasks. To further improve data efficiency, we store only episodes with at least one non-zero reward, or zero-reward episodes in which LTL progression reaches $\top$ or $\bot$, indicating incorrect symbol predictions.

% In the update phase, stored episodes are sampled and each sampled episode's task $\varphi$ is converted into a Moore machine used to initialize an NRM: $NRM_\varphi$.

In the update phase, each sampled episode's task $\varphi$ is converted into a Moore machine and is used to initialize an NRM, denoted by $NRM_\varphi$. The shared grounder is trained by retracing the episodes through the NRM and minimizing the cross-entropy between predicted rewards $\tilde{\boldsymbol{r}} = NRM_\varphi(\boldsymbol{s})$ and ground-truth rewards $\boldsymbol{r}$:
\[
L(\boldsymbol{s}, \boldsymbol{r}) = \text{cross-entropy}(\tilde{\boldsymbol{r}}, \boldsymbol{r}).
\]
Thanks to the differentiability of NRMs, gradients are propagated through time providing a supervision signal to the grounder.

%\paragraph{Example: semisupervised symbol grounding}
%\ELE{TODO}
\subsection{LTL Module}
The final module of our system is the LTL module, which learns an embedding of task formulae and their progressions. In \citet{ltl2action}, Graph Neural Networks (GNNs) proved most effective, with two training schemes: 1) end-to-end training with RL and environment modules, and 2) pretraining via LTL-Bootcamp followed by end-to-end training.  

In LTL-Bootcamp, the LTL and RL modules are trained on a simple sequence-generation task where actions correspond to formula symbols ($A=P$) and the selected symbol is directly observed by the agent. Pretraining imbues the LTL module with task semantics before environment interaction.  
We tested both schemes from \citet{ltl2action} and a third variant using only the bootcamp. The latter was most effective in our sub-symbolic setup, because a fixed LTL module stabilizes learning in the presence of the uncertain grounder. More details can be found in Appendix \ref{sec:effect_pretraining}.

For architecture, we used a Relational Graph Convolutional Network (R-GCN) built from the formula’s Abstract Syntax Tree (AST), with nodes for propositional symbols, logical connectives, or temporal operators, as in \citet{ltl2action} (Figure \ref{fig:framework}c).

\section{Experiments}
% TODO add in main text:
% - plot: online ours vs baseline vs known grounder (and vs symbolic obs?)
% - plot: grounder loss and accuracy
% - generalization table: online ours vs baseline vs known grounder (and vs symbolic obs?), sampling only if needed
% - give consistent names to compared models
We investigate the effectiveness of our approach by comparing it against the one proposed by \citet{encode-ltl-with-rnns}, which serves as our baseline. This method learns a set of sub-networks, one for each proposition and LTL operator, and combines them to form task-specific networks.
While this approach differs substantially from more recent multi-task methods \citep{ltl2action,deepltl}, to the best of our knowledge it is the only previous method for multi-task RL that assumes no knowledge of the labeling function and operates on image observations.
For completeness, we also compare against the upper bound of our method, which replaces NRMs with the true symbol grounding. In our case, this is equivalent to the original LTL2Action \citep{ltl2action}. We compare these methods on two environments, a discrete Minecraft-like environment and a continuous FlatWorld environment. We train using Proximal Policy Optimization (PPO) \citep{ppo} on two different sets of formulae, similar to \citet{ltl2action}. Hyperparameters are reported in Table in Appendix \ref{sec:appendix_hyper}. We evaluate the zero-shot transfer both to unseen formulae from training distributions and to more complex formulae.

\paragraph{Environments}
\label{sec:env}
The \textbf{Minecraft-like} environment is a simple discrete grid-world with 4-way movement actions that contains a variety of items. Each primitive symbol in this environment corresponds to visiting the cell in which a specific item resides. Although similar to the Minecraft Gridworld environment from \citet{neuralrewardmachines}, we use a bigger grid, and the layout of items in the grid-world is randomly sampled at every episode instead of being fixed. In fact, a fixed layout would make learning the symbol grounding as trivial as recognizing one image for each symbol. We note that this also makes the policy itself harder to learn, as the agent cannot rely on a symbol always being associated with a fixed location. The \textbf{FlatWorld} environment \citep{voloshin2023eventualdiscountingtemporallogic} is a continuous 2D plane with continuous movement actions, where primitive symbols correspond to colored circular zones. The placement of the zones is randomly sampled at every episode. Differently to other works, we make sure the zones do not overlap, as to maintain the Declare assumption. More details are provided in Appendix \ref{sec:appendix_envs}.

\paragraph{Tasks}
\label{sec:tasks}
Similar to \citet{ltl2action}, we train on two kinds of formulae: Partially-Ordered tasks, which are conjunctions of sequences that can be executed in parallel, and Global Avoidance tasks, which additionally require avoiding a certain atom for the entire task duration. The automata are required only for training and since translating formulae into automata is too computationally expensive to be done during training, we use a fixed training set of 10.000 formulae whose automata are precomputed. The generalization tasks employ either more sequences \textit{(+conj.)} or longer sequences \textit{(+dep.)}. More details are provided in Appendix \ref{sec:appendix_tasks}.

% This process however remains too computationally expensive to be done online during training, for the tasks considered in our experiments. To solve this problem we create for each class of tasks a fixed train dataset of 10.000 formulae sampled from the original task distributions and compute offline their automata. During training the environments sample from this approximated distribution of tasks instead of using the original sampler function and when training the grounder module we read the corresponding automata instead of computing them.

% all tasks are generated following the syntactically co-safe LTL fragment \citep{optimal-policy-generation-for-partially}

\subsection{Results}

\subsubsection{Minecraft-like Environment}

\begin{figure*}[t]
    \centering
    \begin{subfigure}[T]{0.4925\textwidth}
        \centering
        \caption{Partially-Ordered tasks (Minecraft-like Env.)}
        \includegraphics[width=\textwidth]{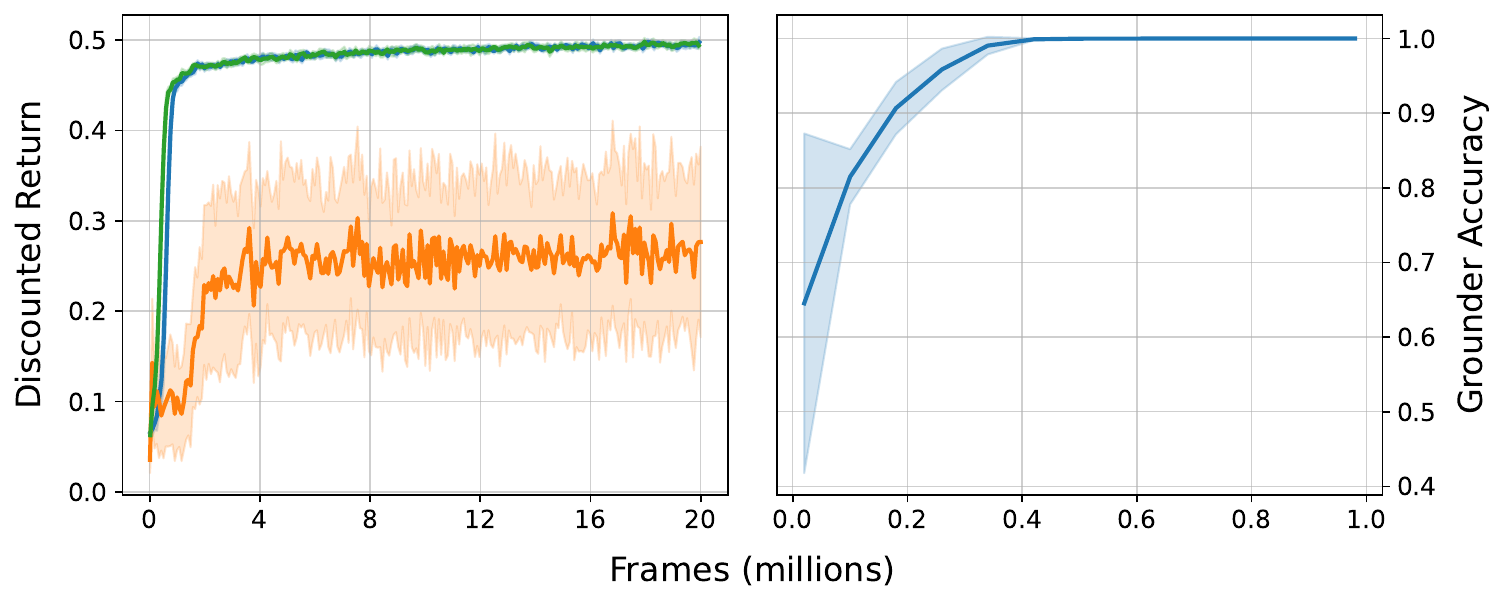}
        \label{fig:gridworld_e_plots}
    \end{subfigure}
    \hfill
    \vspace{-1em}
    \begin{subfigure}[T]{0.4925\textwidth}
        \centering
        \caption{Global Avoidance tasks (Minecraft-like Env.)}
        \includegraphics[width=\textwidth]{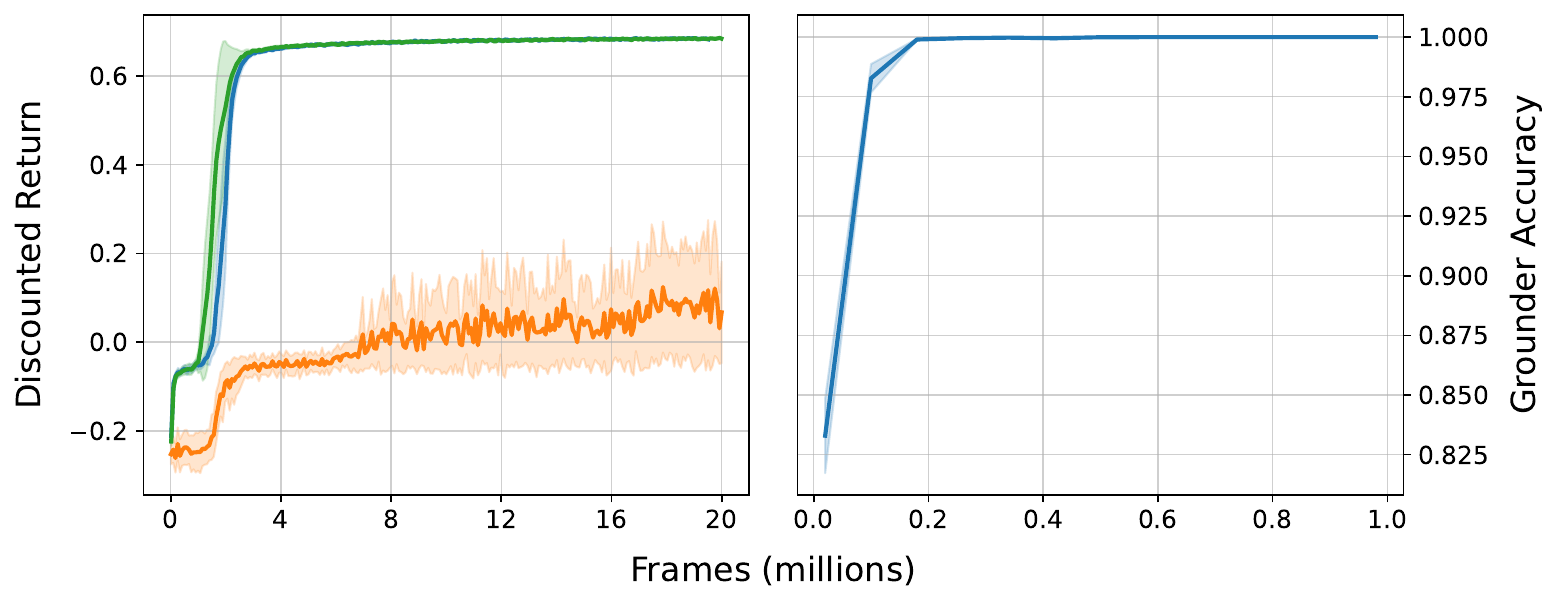}
        \label{fig:gridworld_ga_plots}
    \end{subfigure}
    \begin{subfigure}[T]{0.4925\textwidth}
        \centering
        \caption{Partially-Ordered tasks (FlatWorld Env.)}
        \includegraphics[width=\textwidth]{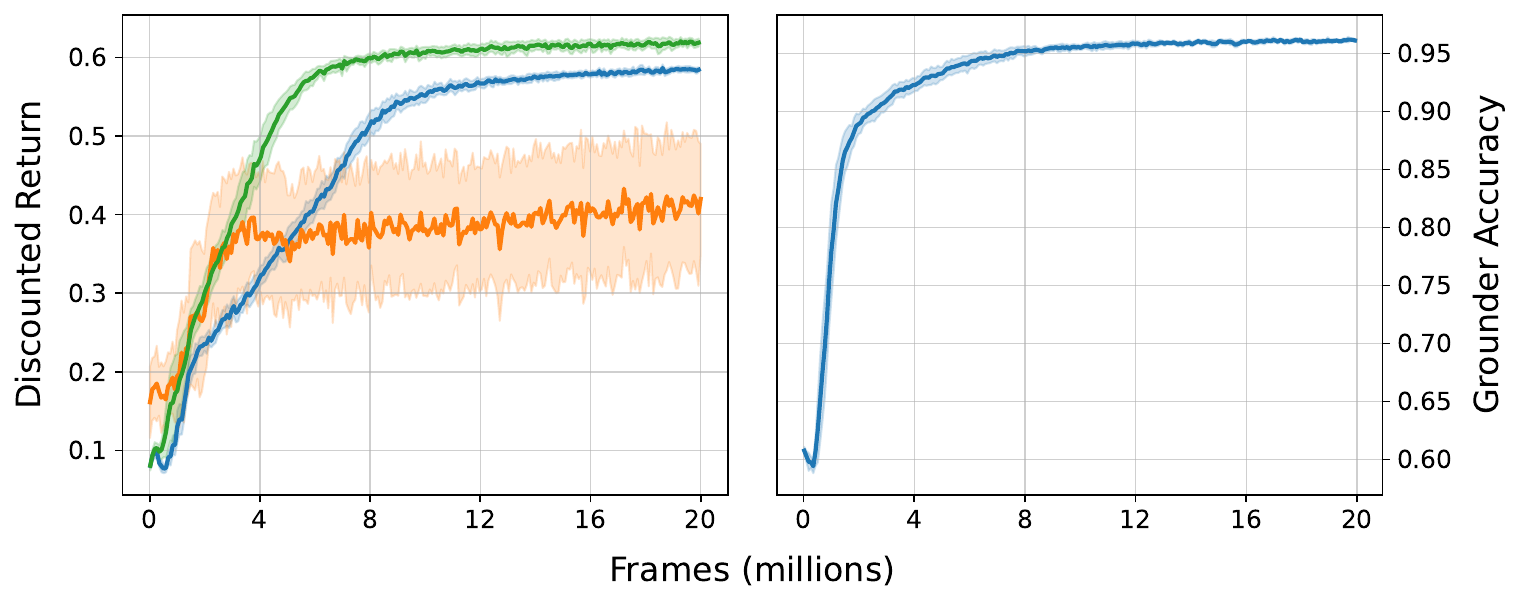}
        \label{fig:flatworld_e_plots}
    \end{subfigure}
    \hfill
    \vspace{-1em}
    \begin{subfigure}[T]{0.4925\textwidth}
        \centering
        \caption{Global Avoidance tasks (FlatWorld Env.)}
        \includegraphics[width=\textwidth]{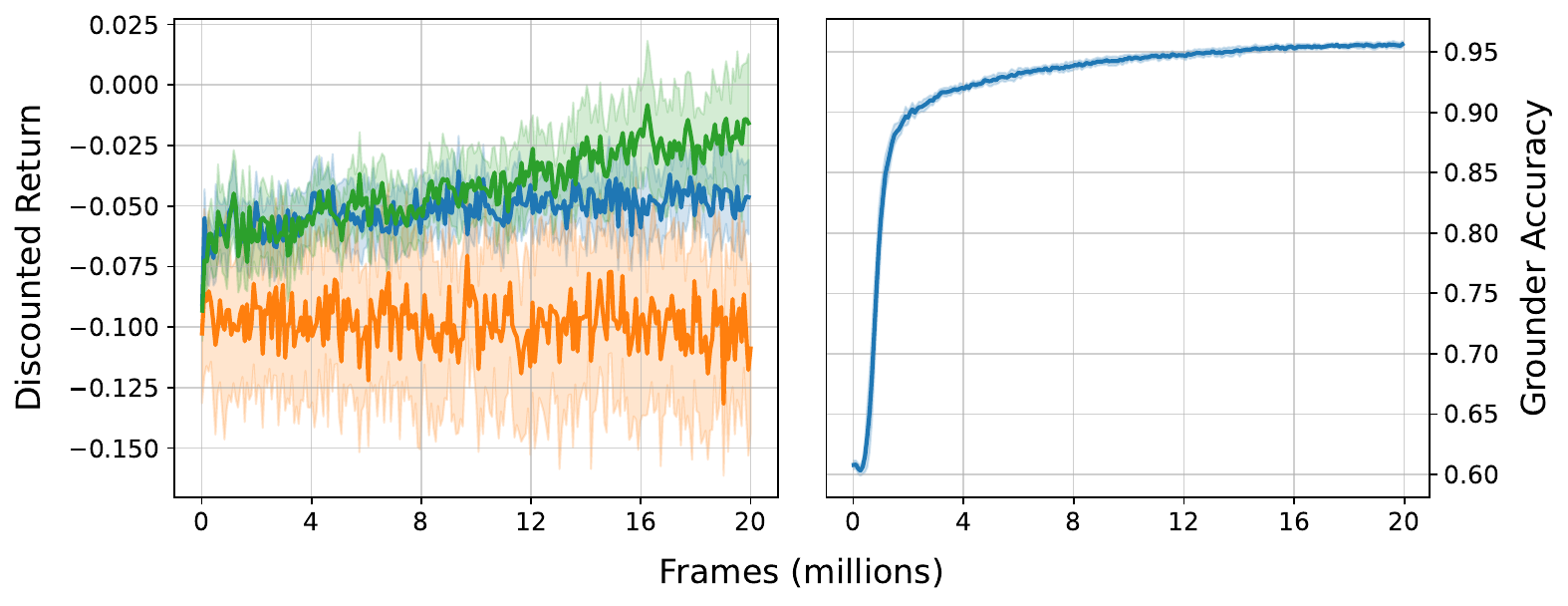}
        \label{fig:flatworld_ga_plots}
    \end{subfigure}
        \begin{subfigure}[T]{0.375\textwidth}
        \centering
        \includegraphics[width=\textwidth]{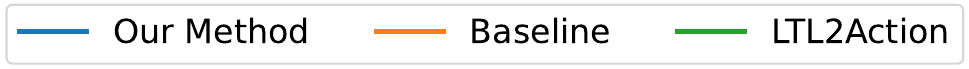}
    \end{subfigure}
    \caption{Comparison between our method (in blue), the baseline (in orange) and LTL2Action (i.e. known symbol grounding) (in green). We report the evolution of the \textit{discounted return} and, for our method, the \textit{grounder accuracy} (averaged over 5 seeds, with std error bands).}
    \label{fig:plots}
\end{figure*}

\begin{table}[t]
\centering
\caption{Average total return and average discounted return (in parentheses) of the RL agents on the Minecraft-like environment (averaged over 5 seeds and 1000 episodes per seed).}
\begin{tabular}{l|cc|c}
\toprule
\multicolumn{4}{c}{Partially-Ordered (Minecraft-like)} \\
\midrule
& Our Method & Baseline & Upper Bound \\
\midrule
Base & \textbf{1.000 (0.488)} & 0.951 (0.170) & 1.000 (0.486) \\
+dep. & \textbf{0.999 (0.039)} & 0.430 (0.001) & 0.999 (0.038) \\
+conj. & \textbf{1.000 (0.194)} & 0.842 (0.048) & 1.000 (0.194) \\
\midrule
\multicolumn{4}{c}{Global Avoidance (Minecraft-like)} \\
\midrule
& Our Method & Baseline & Upper Bound \\
\midrule
Base & \textbf{0.990 (0.680)} & 0.031 (0.022) & 0.992 (0.682) \\
+dep. & \textbf{0.767 (0.223)} & -0.016 (-0.012) & 0.745 (0.214) \\
+conj. & \textbf{0.962 (0.508)} & -0.012 (-0.009) & 0.964 (0.513) \\
\bottomrule
\end{tabular}
\label{tab:upward-generalization-gridworld}
\end{table}

\begin{table}[t]
\centering
\caption{Average total return and average discounted return (in parentheses) of the RL agents on the FlatWorld environment (averaged over 5 seeds and 1000 episodes per seed).}
\begin{tabular}{l|cc|c}
\toprule
\multicolumn{4}{c}{Partially-Ordered (FlatWorld)} \\
\midrule
& Our Method & Baseline & Upper Bound \\
\midrule
Base & \textbf{0.999 (0.610)} & 0.903 (0.311) & 0.999 (0.618) \\
+dep. & \textbf{0.991 (0.322)} & 0.805 (0.072) & 0.998 (0.336) \\
+conj. & \textbf{0.930 (0.403)} & 0.138 (0.018) & 0.983 (0.441) \\
\midrule
\multicolumn{4}{c}{Global Avoidance (FlatWorld)} \\
\midrule
& Our Method & Baseline & Upper Bound \\
\midrule
Base & \textbf{-0.008 (-0.059)} & -0.012 (-0.082) & 0.002 (-0.011) \\
+dep. & \textbf{-0.207 (-0.134)} & -0.250 (-0.162) & -0.175 (-0.118) \\
+conj. & \textbf{-0.185 (-0.122)} & -0.299 (-0.165) & -0.152 (-0.106) \\
\bottomrule
\end{tabular}
\label{tab:upward-generalization-flatworld}
\end{table}

Figures \ref{fig:gridworld_e_plots} and \ref{fig:gridworld_ga_plots} report the training curves of the considered methods and the convergence curve for the grounder accuracy, while Table \ref{tab:upward-generalization-gridworld} reports the final evaluation on both the training formulae and the zero-shot generalization to more complex, unseen formulae.

Overall, our method shows both faster and more stable convergence and significantly higher performances with respect to the baseline, both in terms of success rate and steps needed to solve each formula. In particular, we note that our tasks are composed of much longer sequences with respect to the tasks considered in \citet{encode-ltl-with-rnns}, highlighting how a compositional approach is much more vulnerable to longer tasks. Despite having to learn the symbol grounding online in addition to the policy, our method shows convergence which is very close to the upper bound of training with a known grounder. The same is true for zero-shot generalization to more complex formulae.

The grounder module itself consistently converges within 1 million frames (5\% of a full training run). We observe that the grounder shows faster and more consistent convergence on global avoidance tasks. This is expected as global avoidance tasks are more likely to provide a learning signal to the grounder through the negative reward resulting from formula falsification. On the other hand, partially-ordered tasks only provide a learning signal when the entire formula is satisfied.

\subsubsection{FlatWorld Environment}

Figures \ref{fig:flatworld_e_plots} and \ref{fig:flatworld_ga_plots} show the training curves of the considered methods together with the grounder accuracy, while Table \ref{tab:upward-generalization-flatworld} reports the evaluation on the training formulae and the zero-shot generalization to more complex formulae.

For the partially-ordered tasks, the grounder doesn't reach 100\% accuracy, as in this environment it's harder to determine whether the agent is close enough to a zone to activate its proposition. Despite this, the grounder surpasses 95\% accuracy and the agent achieves performances comparable to the known-grounder scenario, significantly outperforming the baseline, which again fails to scale to tasks much longer than the ones considered in \citet{encode-ltl-with-rnns}.
In contrast, all methods fail to learn the global avoidance tasks, although LTL2Action exhibits a slight improvement over the training period. Despite the failure at the policy level, our method still manages to learn the labeling function and the grounder module surpasses 95\% accuracy.
Regarding zero-shot generalization, our method performs comparably to the known-grounder setting and significantly better than the baseline for the partially-ordered tasks, while it fails for global avoidance tasks, as none of the models are able to successfully learn them in this setting.

\section{Conclusions}

We have presented a method that jointly learns (i) a policy capable of satisfying multiple LTL formulae and (ii) a mapping from raw observations to the atomic propositions appearing in such formulae. Our approach shows little to no performance loss with respect to the upper bound of having access to the true symbol grounder, both on training formulae and on more complex unseen formulae. It also significantly improves over the only previous method addressing the same problem setting of learning multi-task policies without a known symbol grounding. By removing the unrealistic assumption that the true symbol grounding is available, our approach extends multi-task RL to sub-symbolic environments. We believe this represents an important step toward agents capable of following diverse, including previously unseen, instructions directly from raw observations.

%% Future Works

% While the presented results are promising, several aspects remain to be addressed in future work. One aspect is the robustness of the symbol grounder to raw observations, with a basic example being noisy observations. A more realistic setting could also include symbols that correspond to very different images, such as multiple items that can be used for the same purpose. Including symbols that have more varied semantics to being in a specific location would also improve the overall realism.
While the presented results are promising, several directions remain for future work. First, the robustness of the symbol grounder could be improved, for instance by providing noisy observations or considering symbols whose semantics extend beyond the agent's location.
% Another direction for future work is leveraging reward shaping informed by the task progression information, which could highly benefit the initial learning phase where neither the policy nor the grounder can produce meaningful behaviors. For the same reason, introducing some form of curriculum learning, where easier formulae are trained first, could also be beneficial. 
Second, learning could benefit from reward shaping informed by task progression, particularly during early training when neither the policy nor the grounder produces meaningful behavior. For the same reason, curriculum learning, where the agent is trained on increasingly complex formulae, could also be beneficial.
% Finally, building on recent work on fuzzy logic conformance checking \citep{fuzzy-logic,fuzzy-logic2} could in principle allow to train the symbol grounder without explicitly constructing the automaton corresponding to each LTL formula.
Finally, building on recent work on fuzzy logic conformance checking \citep{fuzzy-logic,fuzzy-logic2} could allow to train the symbol grounder without explicitly constructing the automata for the LTL formulae.

\appendix

%%%%%%%%%%%%%%%%%%%%%%%%%%%%%%%%%%%%%%%%%%%%%%%%%%%%%%%%%%%%%%%%
%% NOTE: THIS MARKS THE END OF THE "MAIN TEXT"
%%%%%%%%%%%%%%%%%%%%%%%%%%%%%%%%%%%%%%%%%%%%%%%%%%%%%%%%%%%%%%%%

%%%%%%%%%%%%%%%%%%%%%%%%%%%%%%%%%%%%%%%%%%%%%%%%%%%%%%%%%%%%%%%%
%% Bibliography
%%%%%%%%%%%%%%%%%%%%%%%%%%%%%%%%%%%%%%%%%%%%%%%%%%%%%%%%%%%%%%%%
\bibliography{main}
\bibliographystyle{rlj}

%%%%%%%%%%%%%%%%%%%%%%%%%%%%%%%%%%%%%%%%%%%%%%%%%%%%%%%%%%%%%%%%
% AUTHOR: If your paper has no supplementary materials, you may 
%         comment out the line below, which creates the title for
%         the supplementary materials.
%%%%%%%%%%%%%%%%%%%%%%%%%%%%%%%%%%%%%%%%%%%%%%%%%%%%%%%%%%%%%%%%
\beginSupplementaryMaterials

\section{Experimental Details}

\subsection{Tasks}
\label{sec:appendix_tasks}

The set of LTL goals used for our experiments are generated following the syntactically co-safe LTL fragment, ensuring they are co-safe LTL formulae.

\subsubsection{Partially-Ordered Tasks}

Partially-Ordered tasks consist of multiple sequences of propositions that can be satisfied in parallel: while the propositions within each sequence must be achieved in order, the different sequences can be interleaved or solved simultaneously. They can be described with production rules of a context-free grammar:
\begin{align*}
    \text{formula} & \;\; \rightarrow \;\; \text{sequence} \; | \; (\text{formula} \wedge \text{sequence} ) \\
    \text{sequence} & \;\; \rightarrow \;\; \LTLeventually \text{term} \; | \; \LTLeventually (\text{term} \wedge \text{sequence} ) \\
    \text{term} & \;\; \rightarrow \;\; \text{prop} \; | \; \text{prop} \vee \text{prop}
\end{align*}
Where \textit{prop} is a propositional atom from the environment’s set of propositions $\mathcal{P}$. The probability of a term being replaced with a disjunction of two atoms is of 25\%.

The Minecraft-like environment uses tasks whose number of sequences is sampled between 1 and 4, while the length of each sequence is sampled between 1 and 5. The upward generalization tasks are: (i) tasks with increased depth from a maximum of 5 to a fixed 15 and (ii) tasks with increased number of conjunctions from a maximum of 4 to a fixed 12. The FlatWorld environment instead employs a single sequence of length between 1 and 3. The upward generalization tasks are: (i) tasks with increased depth from a maximum of 3 to a fixed 4 and (ii) tasks with increased number of conjunctions from a maximum of 1 to a fixed 2.

\subsubsection{Global Avoidance Tasks}

True global avoidance cannot be expressed as a co-safe property. However the construct $\neg a \LTLuntil b$ can be concatenated to create sequential tasks where between two steps a certain proposition has to be avoided. The context-free grammar describing the conjunction of such sequences is:
\begin{align*}
    \text{formula} \;\; \rightarrow \;\; & \text{sequence} \; | \; (\text{formula} \wedge \text{sequence} ) \\
    \text{sequence} \;\; \rightarrow \;\; & (\text{avoidance} \; \LTLuntil \; \text{prop}) \; | \\
    & (\text{avoidance} \; \LTLuntil \; ( \text{prop} \wedge \text{sequence}) ) \\
    \text{avoidance} \;\; \rightarrow \;\; & \neg \text{prop}
\end{align*}
With the additional constraint of having the \textit{avoidance} rule always using the same proposition, which instead cannot be used in the rest of the rules, we create a co-safe task where a certain proposition must be avoided for the duration of the task, but as soon as the sequential task is completed the avoidance is lifted and the formula is satisfied.

The Minecraft-like environment uses tasks whose number of sequences is sampled between 1 and 2, while the length of each sequence is sampled between 1 and 3. The upward generalization tasks are: (i) tasks with increased depth from a maximum of 3 to a fixed 5 and (ii) tasks with increased number of conjunctions from a maximum of 2 to a fixed 3. 
The FlatWorld environment instead employs a single sequence of length between 1 and 2. The upward generalization tasks are: (i) tasks with increased depth from a maximum of 2 to a fixed 3 and (ii) tasks with increased number of conjunctions from a maximum of 1 to a fixed 2.

\subsection{Environments}
\label{sec:appendix_envs}

\subsubsection{Minecraft-Like Environment}

The main environment we consider for our experiments is a variation of the classic GridWorld environment. It consists of a discrete $7\times7$ grid, where each cell corresponds to a state of the environment and is associated with at most one proposition from a set of atomic propositions $\mathcal{P}$. In each cell it's possible to move along the four cardinal directions and if the agent’s move would take it outside the grid, then the position wraps around to the opposite side of the grid along that axis.

For our experiments we consider a set of 5 atomic propositions $\mathcal{P} = \{ a,b,c,d,e \}$, each associated with a visual representation, as illustrated in Table \ref{tab:propositions}. In particular each proposition is linked to two cells in the grid and the agent always starts in a cell without propositions. Furthermore, the symbols' locations and the agent's starting position are randomized each episode.

\begin{table}[h]
\centering
\caption{List of the atomic propositions $\mathcal{P}$ with their name, symbolic representation for the tasks, numerical identifier and icon representation for image observations.}
\begin{tabular}{c c >{\centering\arraybackslash}m{1.0cm} >{\centering\arraybackslash}m{1.5cm}}
\toprule
\multicolumn{4}{c}{Propositions $\mathcal{P}$} \\
\midrule
Name & Symbol & Number & Icon \\
\midrule
pick & $a$ & $0$ & \includegraphics[width=0.75cm]{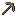} \\
lava & $b$ & $1$ & \includegraphics[width=0.75cm]{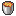} \\
door & $c$ & $2$ & \includegraphics[width=0.75cm]{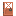} \\
apple  & $d$ & $3$ & \includegraphics[width=0.75cm]{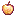} \\
egg  & $e$ & $4$ & \includegraphics[width=0.75cm]{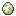} \\
\bottomrule
\end{tabular}
\label{tab:propositions}
\end{table}

\begin{figure}[h]
  \centering
  \begin{subfigure}[t]{0.35\linewidth}
    \centering
    \includegraphics[width=\linewidth]{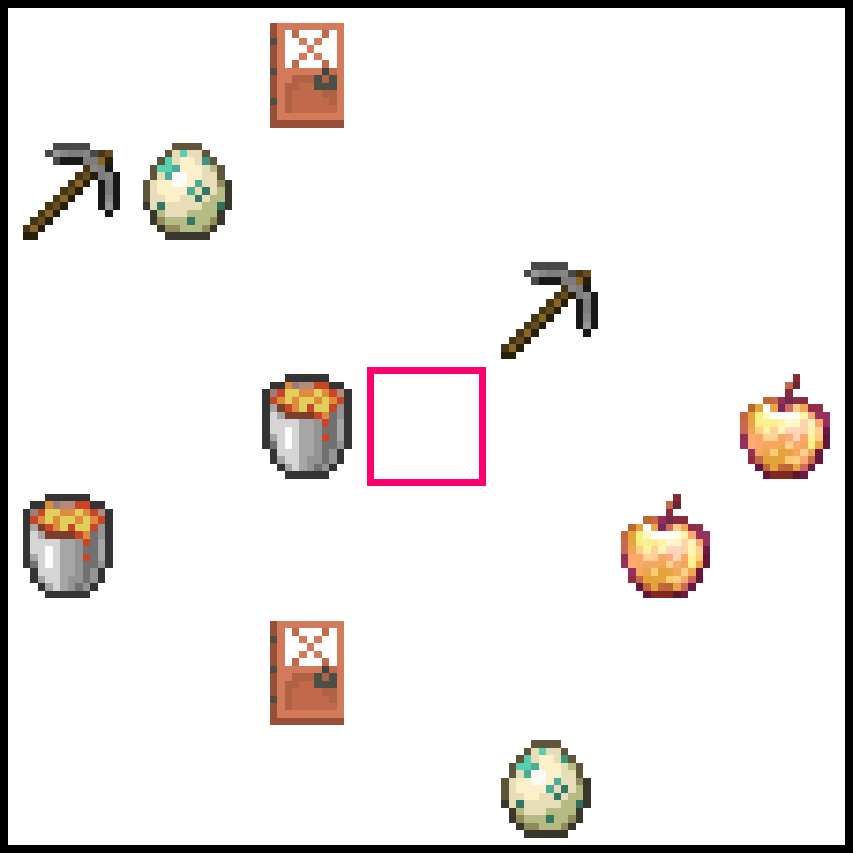}
    \caption{Native Observation}
    \label{fig:example_environment_minecraft}
  \end{subfigure}
  \hspace{0.05\textwidth}
  \begin{subfigure}[t]{0.35\linewidth}
    \centering
    \includegraphics[width=\linewidth]{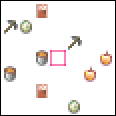}
    \caption{$56 \times 56$ Observation}
    \label{fig:example_observation_minecraft}
  \end{subfigure}
  \caption{Example of the image observations of the \textit{Minecraft-like environment} following the \textit{egocentric view}. The agent position is represented by the red square icon.}
  \label{fig:example_env}
\end{figure}

The observations are RGB images of size $56 \times 56$ representing the map visually as a $7 \times 7$ grid of $8 \times 8$ versions of the propositions' visual representations. The resolution of the observations provided to the agent is lower than the native resolution, as shown in Figure \ref{fig:example_env}. Furthermore, the observations are shifted in order to represent an \textit{egocentric view} of the agent, with the cell representing the current position shifted to the center of the representation.

\subsubsection{FlatWorld}

This environment is a 2D continuous world, with a continuous velocity action space. The atomic propositions are associated to circular colored regions within the map and these regions cannot overlap, in order to maintain the Declare assumption. The colored regions' location are randomized each episode and the agent's starting position is sampled from the space where no proposition hold.

\begin{figure}[h]
  \centering
  \begin{subfigure}[t]{0.35\linewidth}
    \centering
    \includegraphics[width=\linewidth]{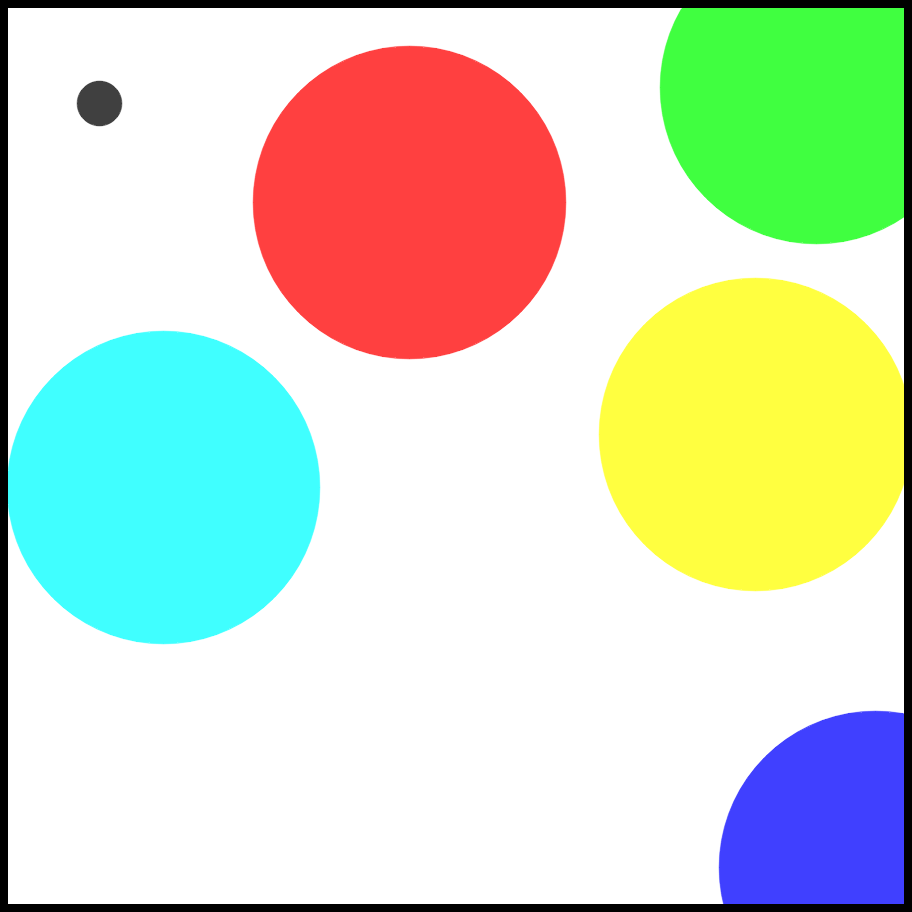}
    \caption{Native Observation}
    \label{fig:example_environment_flatworld}
  \end{subfigure}
  \hspace{0.05\textwidth}
  \begin{subfigure}[t]{0.35\linewidth}
    \centering
    \includegraphics[width=\linewidth]{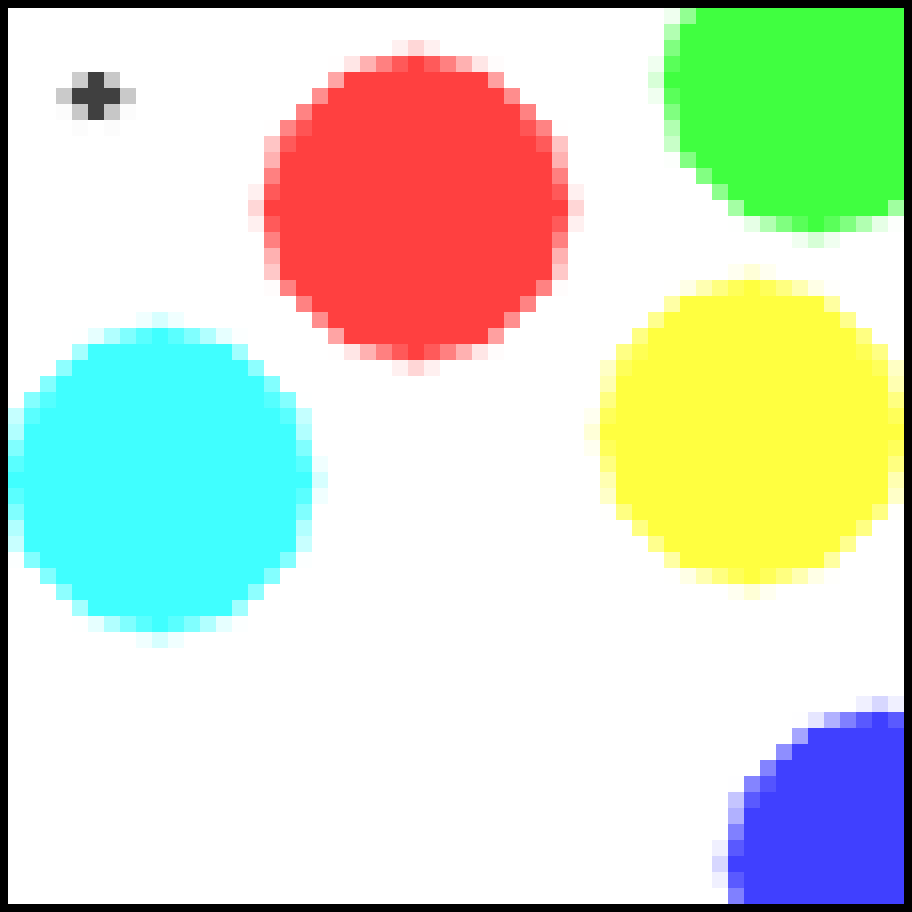}
    \caption{$56 \times 56$ Observation}
    \label{fig:example_observation_flatworld}
  \end{subfigure}
  \caption{Example of the image observations of the \textit{FlatWorld environment}. The agent position is represented by the black dot.}
  \label{fig:example_env_flatworld}
\end{figure}

\subsection{Hyperparameters} \label{sec:appendix_hyper}

The hyperparameters used for training our architecture are reported in Tables \ref{tab:hyperparameters} and \ref{tab:grounder_hyperparameters}. The hyperparameters used for the baseline are reported in Table \ref{tab:baseline_hyperparameters}, while the upper bound with known labeling function (equivalent to LTL2Action on the image-based environment) employ those of Table \ref{tab:hyperparameters}.

\begin{table}[h]
\centering
\caption{Hyperparameter settings for the agent training algorithm used in the LTL module pretraining and the complete trainings. The same set of hyperparameters are used for Partially-Ordered and Global Avoidance task classes.}
\begin{tabular}{l|ccc}
\toprule
\multicolumn{4}{c}{Agent Training Hyperparameters} \\
\midrule
& Bootcamp & GridWorld & FlatWorld \\
\midrule
Steps per update & 8192 & 2048 & 2048 \\
N. epochs & 2 & 4 & 4 \\
Minibatch size & 1024 & 256 & 256 \\
Discount factor & 0.9 & 0.94 & 0.94 \\
Learning rate & 0.001 & 0.0003 & 0.0003 \\
GAE-$\lambda$ & 0.5 & 0.95 & 0.95 \\
Entropy coeff. & 0.01 & 0.01 & 0.01 \\
Value loss coeff. & 0.5 & 0.5 & 0.5 \\
Gradient clip & 0.5 & 0.5 & 0.5 \\
PPO clip & 0.1 & 0.2 & 0.2 \\
\bottomrule
\end{tabular}
\label{tab:hyperparameters}
\end{table}

\begin{table}[h]
\centering
\caption{Hyperparameters for the grounder training algorithm.}
\begin{tabular}{l|cc}
\toprule
\multicolumn{3}{c}{Grounder Training Hyperparameters} \\
\midrule
& GridWorld & FlatWorld \\
\midrule
Learning rate & 0.001 & 0.001 \\
Buffer size & 2048 & 8192 \\
Val. Buffer size & 512 & 2048 \\
Batch size & 16 & 16 \\
Accumulation & 4 & 8 \\
Update steps & 64 & 128 \\
Patience & 250 & 4000 \\
\bottomrule
\end{tabular}
\label{tab:grounder_hyperparameters}
\end{table}

\begin{table}[h]
\centering
\caption{Hyperparameter settings for the baseline agent training algorithm. The same set of hyperparameters are used for Partially-Ordered and Global Avoidance task classes.}
\begin{tabular}{l|cc}
\toprule
\multicolumn{3}{c}{Baseline Training Hyperparameters} \\
\midrule
& GridWorld & FlatWorld \\
\midrule
Steps per update & 1024 & 1024 \\
N. epochs & 1 & 1 \\
Minibatch size & 256 & 256 \\
Discount factor & 0.94 & 0.94 \\
Learning rate & 0.0003 & 0.0003 \\
GAE-$\lambda$ & 0.95 & 0.95 \\
Entropy coeff. & 0.001 & 0.001 \\
Value loss coeff. & 0.5 & 0.5 \\
Gradient clip & 0.5 & 0.5 \\
PPO clip & 0.2 & 0.2 \\
\bottomrule
\end{tabular}
\label{tab:baseline_hyperparameters}
\end{table}

\subsection{Effects of Pretraining} \label{sec:effect_pretraining}

\begin{figure}
    \centering
    \includegraphics[width=0.75\textwidth]{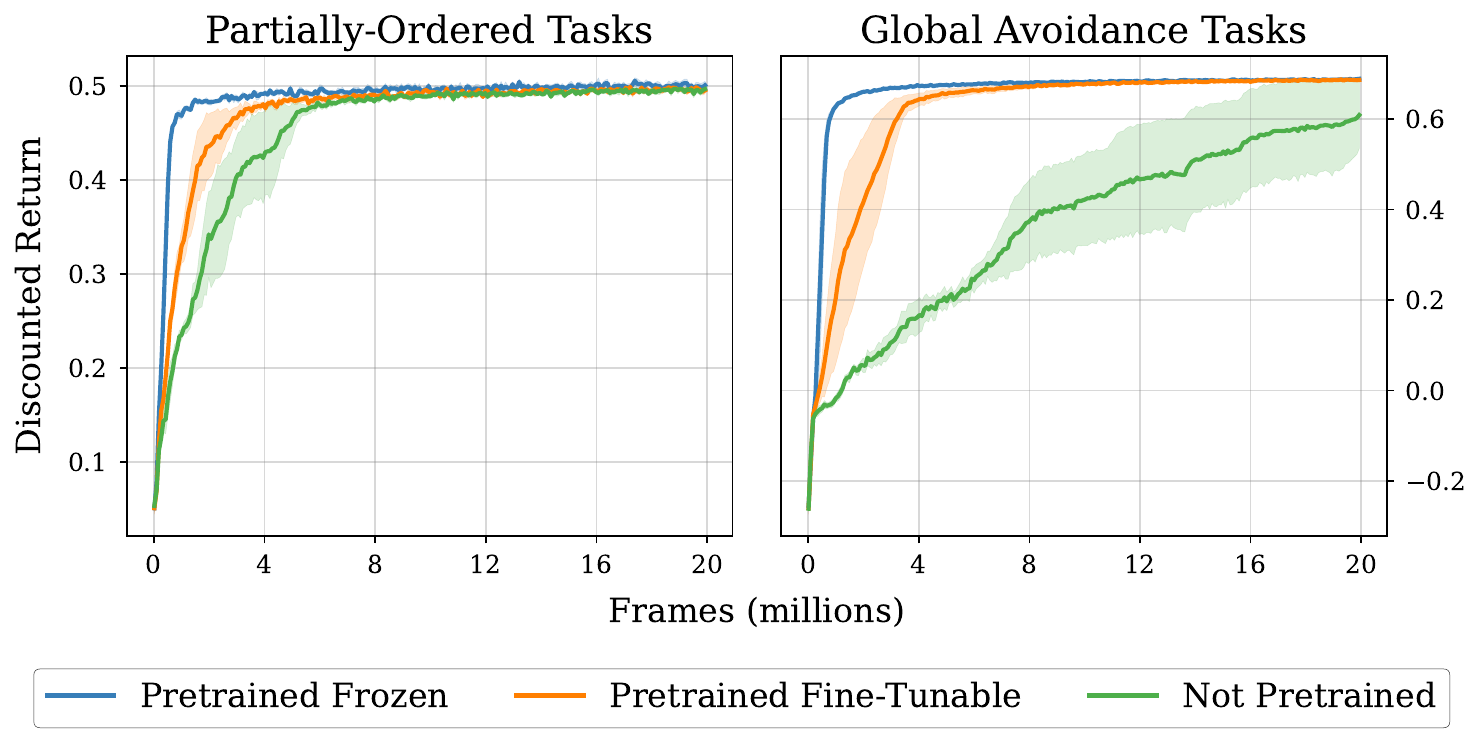}
    \caption{Comparison between the 3 pretraining settings: non-pretrained LTL module (in green), pretrained fine-tunable LTL module (in orange) and pretrained frozen LTL module (in blue). We report the evolution of the \textit{discounted return} during the training (averaged over 5 seeds, with standard deviation error bands).}
    \label{fig:effects_of_pretrain}
\end{figure}

We investigate the effects of the LTL module pretraining on the downstream task learning process. The pretraining is performed on the LTLBootcamp environment for $10$ million frames using the pretraining hyperparameters from Table \ref{tab:hyperparameters} for both Partially-Ordered and Global Avoidance tasks. The pretrained LTL module is then transferred into a new agent, which is then trained on the Minecraft-like environment (with known labeling function) for 20 million frames following the hyperparameters from Table \ref{tab:hyperparameters}. We compare 3 settings: (1) non-pretrained LTL module, (2) pretrained and fine-tunable LTL module and (3) pretrained and frozen LTL module. As shown in Figure \ref{fig:effects_of_pretrain}, using the pretrained LTL module, both frozen and fine-tunable, results in significantly faster convergence and therefore a higher sampling efficiency. This is particularly evident in the Global Avoidance tasks, where the non-pretrained agents show higher variability during training and not all runs manage to reach convergence within the $20$ million frames, while all the pretrained agents are close to convergence within the first $5$ million frames. This could denote that, despite being significantly shorter, the Global Avoidance tasks are inherently harder to solve, because of the possibility of failure. Focusing on the pretraining strategies, using a pretrained and frozen LTL module performs better in both settings. An intuition motivating this is that during the first training steps of the fine-tunable agent, when the other modules have only been initialized, the backpropagation partially disrupts the pretraining.

\begin{figure}
    \centering
    \includegraphics[width=0.75\textwidth]{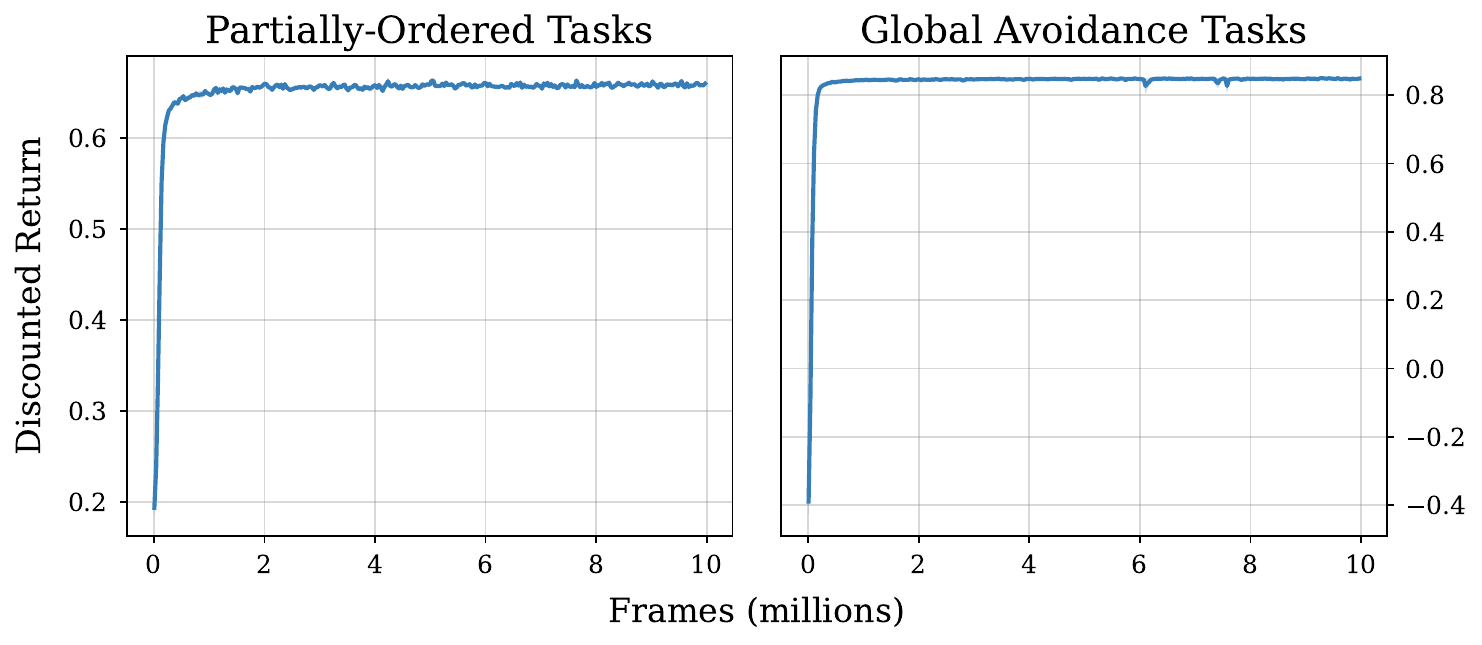}
    \caption{Evolution of the \textit{discounted return} obtained by the dummy agent over the duration of the pretraining on the LTLBootcamp environment (averaged over 5 seeds, with standard deviation error bands).}
    \label{fig:pretrain}
\end{figure}

The plots in Figure \ref{fig:pretrain} show the learning curves of the pretraining scheme's agent over the LTLBootcamp environment. Compared to the curves in Figure \ref{fig:effects_of_pretrain}, we can see both a faster convergence and a significantly higher discounted return, despite using the same task distributions. This is expected, since the LTLBootcamp environment is built to be a toy environment where the tasks can be solved more easily and in fewer steps, leading to a smaller discount and a less sparse reward. Another important factor that accelerates the training is the simplicity of the RL agent, which does not use any environment module and has a smaller RL module.

\end{document}